\documentclass[runningheads]{llncs}
\usepackage[T1]{fontenc}
\usepackage{graphicx}
\usepackage{hyperref}
\usepackage{color}

\usepackage{relsize}
\usepackage{multirow}
\usepackage{amsmath}
\usepackage{makecell}
\usepackage{tabularx}
\usepackage{marvosym}

\begin{document}
\title{Resolving Variable Respiratory Motion From Unsorted 4D Computed Tomography}
%
\titlerunning{Variable motion from unsorted 4DCT}
%
\author{Yuliang Huang\inst{1,2}\textsuperscript{(\Letter)} \and
Bjoern Eiben\inst{3} \and
Kris Thielemans\inst{1,4} \and
Jamie R. McClelland\inst{1,2}}

%
\authorrunning{Y. Huang et al.}
%
\institute{
Centre for Medical Image Computing, Department of Medical Physics and Biomedical Engineering, University College London, United Kingdom \email{yuliang.huang.21@ucl.ac.uk} \and
Wellcome/EPSRC Centre for Interventional and Surgical Sciences, University College London, London, United Kingdom \and
Joint Department of Physics, The Institute of Cancer Research and The Royal Marsden NHS Foundation Trust, London, United Kingdom \and
Institute of Nuclear Medicine, University College London, London, United Kingdom 
}
\maketitle              
\begin{abstract}
4D Computed Tomography (4DCT) is widely used for many clinical applications such as radiotherapy treatment planning, PET and ventilation imaging. However, common 4DCT methods reconstruct multiple breath cycles  into a single, arbitrary breath cycle which can lead to various artefacts, impacting the downstream clinical applications. Surrogate driven motion models can estimate continuous variable motion across multiple cycles based on CT segments `unsorted' from 4DCT, but it requires respiration surrogate signals with strong correlation to the internal motion, which are not always available. The method proposed in this study eliminates such dependency by adapting the hyper-gradient method to the optimization of surrogate signals as hyper-parameters, while achieving better or comparable performance, as demonstrated on digital phantom simulations and real patient data. Our method produces a high-quality motion-compensated image together with estimates of the motion, including breath-to-breath variability, throughout the image acquisition. Our method has the potential to improve downstream clinical applications, and also enables retrospective analysis of open access 4DCT dataset where no respiration signals are stored. Code is avaibale at \url{https://github.com/Yuliang-Huang/4DCT-irregular-motion}.
\end{abstract}
\section{Introduction}
4D Computed Tomography (4DCT) can capture 3DCT images of each respiration motion state in a breathing cycle \cite{taguchi2003temporal}, thus reducing motion artefacts. The basic principle is to synchronize the CT acquisition with a respiration signal measured by external devices such as  a belt or skin marker. Most commonly, respiratory motion states are defined based on phase, i.e. intervals between adjacent peaks/valleys of the respiration signal are evenly divided into several (normally ten) respiration phases. CT data acquired at different couch positions are then sorted and stacked to form the 3DCT of each respiration phase \cite{ford2003respiration}.

4DCT is used to delineate tumors and critical organs for almost all the thoracic and abdominal cancer patients treated by radiotherapy \cite{muirhead2008use}.  The tendency to increase the use of proton radiotherapy will place a higher emphasis on motion management guided by 4DCT \cite{taasti2021treatment}. 4DCT also has been applied in other domains outside radiotherapy, such as PET-imaging for attenuation correction \cite{bettinardi2010detection}, and ventilation imaging \cite{low2021ventilation} for ablative pulmonary interventions.

\begin{figure}[t]
\includegraphics[width=\textwidth]{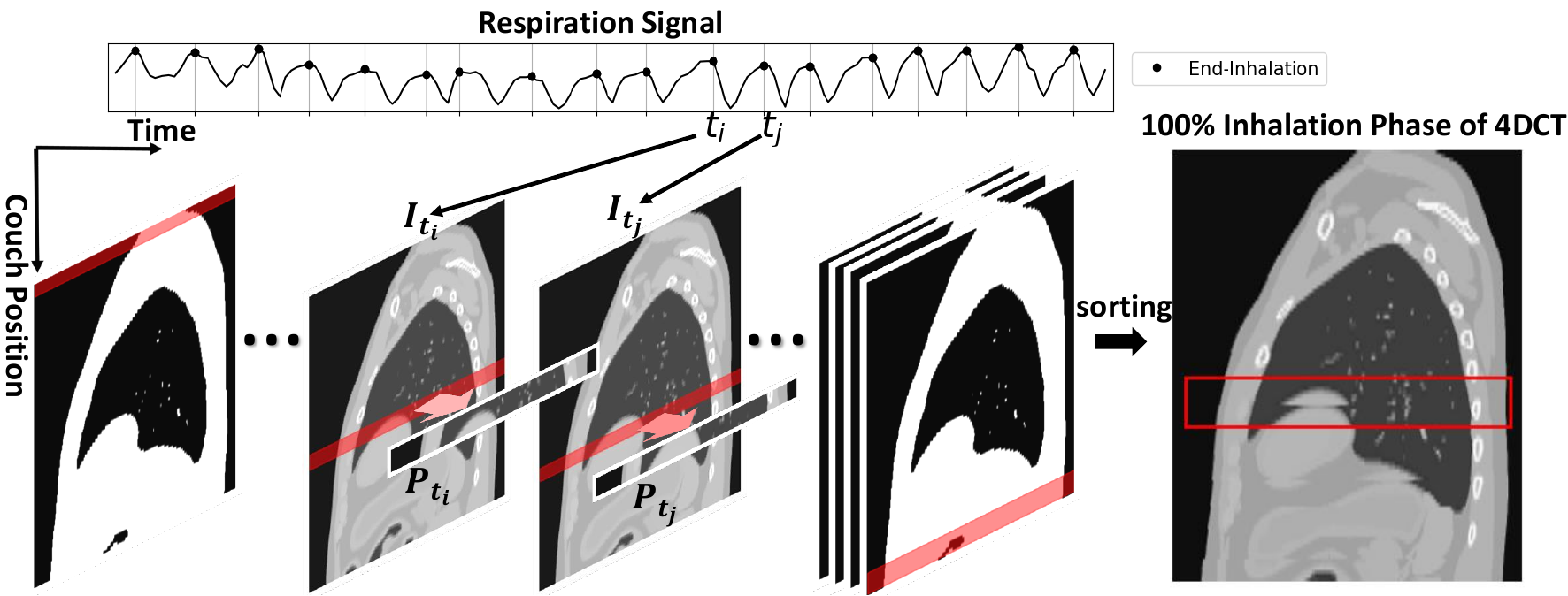}
\caption{ Illustration of 4DCT artefacts caused by irregular breathing. Time points of the end-inhalation phase are identified as black dots in the respiration signal plot, with two of them being $t_i$ and $t_i$. $I_{t_i}$ and $I_{t_j}$ are the real-time volumes at $t_i$ and $t_j$ respectively. $P_{t_i}$ and $P_{t_j}$ indicate the CT segments acquired at the couch positions corresponding to $t_i$ and $t_i$, which are stacked into the final end-inhalation phase image. Artefact of duplicated diaphragm can be observed in the end-inhalation phase image. } \label{fig1}
\end{figure}

Despite its critical role in various applications, a  fundamental challenge to 4DCT is irregular breathing, which has not been solved \cite{tryggestad20234dct}. Fig.~\ref{fig1} illustrates an artefact caused by irregular breathing, using the end-inhalation phase as an example. As shown by Fig.~\ref{fig1}, the patient had deeper inhalation at $t_j$ than $t_i$, causing the diaphragm to be lower at $t_j$ and  appear in the CT segments acquired at the two couch positions. The final end-inhalation images, formed by stacking these CT segments, will therefore show misalignment, duplicate, missing or distortion of structures. Indeed, clinical evidence has shown  correlation between 4DCT motion artefacts and worse control of lung and liver cancer metastasis after radiotherapy \cite{sentker20204d}, indicating the significance of overcoming the irregular breathing problem for 4DCT.

Different solutions have been proposed in the past two decades. Different sorting methods \cite{abdelnour2007phase,li2012clinical,werner2016optimized} could produce fewer artefacts, but might lead to gaps in slice locations for certain phase \cite{abdelnour2007phase}. Other studies were based on pairwise/groupwise registration \cite{zhang2013modeling,li2017automated,shao2021geodesic}, where a motion-free template image could be obtained from all the phase images warped to the template image space, and then used to recover motion-free images of each phase by backward warping. It is unknown how the residual motion artefacts within 4DCT would impact the pairwise/groupwise registration and these methods could only recover arbitrary breath cycle, unable to address errors/uncertainties in the motion seen in the 4DCT. Low \textit{et al.} \cite{low2013novel} developed a 5DCT protocol using a  5D lung motion model \cite{zhao2009characterization} but required a special acquisition protocol. Similar to 5DCT method, McClelland \textit{et al.} \cite{mcclelland2017generalized} proposed the surrogate-driven motion model method that fits in with standard protocol, which could estimate a motion-free reference image and the motion fields at each timepoint based on CT segments. Therefore, phase-binning was not required and the irregular breath would not cause artefact.

Although promising, the surrogate-driven motion model relies on existence of surrogate signals that have strong correlation with internal motion, which are challenging to obtain. Weak or unstable correlation between the surrogate signals and internal motion can significantly impact model precision \cite{huang2024surrogate}. More importantly, respiration signals are often absent from retrospective data, especially open access data, and some hospitals do not keep the respiration signals after 4DCT data are sorted, so the method cannot be applied retrospectively.

In this work, we propose a method that improves the surrogate-driven motion model. The proposed method uses the hypergradient method \cite{baydin2017online} to obtain the suitable surrogate signals when the clinical respiration signals are unavailable (denoted as surrogate-free motion model), or optimize existing surrogate signals so that the correlation with internal motion can be strengthened (denoted as surrogate-optimized motion model). A phantom simulation with ground-truth is used to quantitatively evaluate our methods against other methods in literature. Furthermore, to demonstrate the benefits of our method when no respiration signals are available, we retrospectively process 4DCTs from local cancer center and public available dataset by our method and show that the irregular breathing motion could be estimated by our method.

\section{Methodology}
\subsubsection{Problem Formulation}
During 4DCT scan, only a few slices of the whole volume will be acquired at each time and the couch will move to change the slice locations being imaged. The CT segment $P_t$ at time $t$ can be obtained by
\begin{equation}
P_t = E_{\mathcal{Z}_t}(I_t), \mathcal{Z}_t=[z_{t}^\mathrm{inf},z_{t}^\mathrm{sup}]
\end{equation}
where $I_t$ is the full dynamic volume at time $t$, $E$ is the slice extraction operator and $\mathcal{Z}_t$ is the slice locations being imaged at time $t$.
If there is a reference volume $I_0$, the dynamic volumes could be transformed from $I_0$ as below:
\begin{equation}
I_t = T(I_0, M_t)
\end{equation}
where $T$ is a function that can resample $I_0$ according to given transformation, and $M_t$ is the parameters that determine the transformation from dynamic volume space to reference volume space at time $t$. Here the free-form cubic bspline transformation is adopted, so $M_t$ refers to the control point grid displacement field. 
To get the dynamic volumes at all the $N_t$ timepoints, a motion-free $I_0$ and the transformations over time $\{M_t | t\in [1, N_t]\}$ need to be estimated.

\subsubsection{Surrogate-Driven Motion Model Approach}
Directly solving $M_t$ is impossible given the ill-posedness of the question. The surrogate-driven motion model approach \cite{mcclelland2017generalized} models the transformation parameters as linear combination of products of surrogate signals and spatial correspondence models:
\begin{equation} \label{eq:motionmodel}
   M_t =  \Sigma_{i=1}^{N_s} S_{i,t} C_i
\end{equation}
where $N_s$ is the number of surrogate signals, $S_{i,t}$ is the $i^{th}$ surrogate signal at time $t$, and $C_i$ is the corresponding spatial correspondence model with the same degree of freedom as $M_t$ that indicates major breathing patterns over space. The surrogate signals can be adapted from respiration signals measured by external devices and are fixed during the motion model fitting.

The motion model parameters can be solved by optimizing the similarity such as mean-square-error between the estimated and acquired CT segments over time:

\begin{equation}
    f(C) =  \Sigma_{t=1}^{N_t} || E_{\mathcal{Z}_t}(T(I_0, \Sigma_{i=1}^{N_s} S_{i,t} C_i)) - P_t ||_{2}^{2}
\end{equation}

Gradient descent can be used to minimize the objective function $f$:
\begin{equation}\label{eq:grad1}
    C^{(k)} = C^{(k-1)} - \lambda^{(k)} \Sigma_{t=1}^{N_t} S_{\cdot,t} \cdot \nabla_{M_t}{ f(C^{(k-1)})} 
\end{equation}
where $C^{(k)}$ is the updated motion model parameters after the $k^{th}$ iteration, and $\lambda^{(k)}$ is the learning rate that can be determined by line search.

\subsubsection{Proposed Method: Surrogate-Free/Optimized Motion Model}
The surrogate signals are fixed in the method above. In contrast, we propose to treat the surrogate signals as optimizable hyper-parameters beside the spatial correspondence model. Inspired by hypergradient proposed in \cite{baydin2017online}, we calculate the following hypergradient w.s.t. the surrogate signals through the chain rule:
\begin{equation}\label{eq:hypergrad1}
    h^{(k)}_{i,t} = \frac{ \partial{f(C^{(k-1)})} }{ \partial{S_{i,t}} }= \frac{ \partial{(\Sigma_{i=1}^{N_s}S_{i,t}C_i^{(k-1)}} )}{ \partial{S_{i,t}} } \cdot\nabla_{M_t}{ f(C^{(k-1)})}
\end{equation}

By using the update rule of the previous iteration:
\begin{equation}\label{eq:previous}
C^{(k-1)} = C^{(k-2)} - \lambda^{(k-1)} \Sigma_{t=1}^{N_t} S_{\cdot,t} \cdot \nabla_{M_t}{ f(C^{(k-2)})}
\end{equation}
and assuming that optimal values of $S$ do not change much between two consecutive iterations, we can get
\begin{equation}\label{eq:hypergrad2}
\begin{split}
    \frac{ \partial{(\Sigma_{i=1}^{N_s}S_{i,t}C_i^{(k-1)}} )}{ \partial{S_{i,t}} } &= C_i^{(k-1)} + S_{i,t}\frac{\partial{C_i^{(k-1)}}}{\partial{S_{i,t}}} \\
    &= C_i^{(k-1)} - \lambda^{(k-1)}S_{i,t} \nabla_{M_t}{ f(C^{(k-2)})}
\end{split}
\end{equation}
Note that the second equality is obtained by plugging in Eqn.~\ref{eq:previous}. 

Substituting Eqn.~\ref{eq:hypergrad2} back into Eqn.~\ref{eq:hypergrad1} will yield
\begin{equation}
    h^{(k)}_{i,t} = [C_i^{(k-1)} - \lambda^{(k-1)} S_{i,t}^{(k-1)} \nabla_{M_t}{ f(C^{(k-2)})}] \cdot \nabla_{M_t}{ f(C^{(k-1)})}
\end{equation}
and the surrogate signals can then be updated accordingly:
\begin{equation}
    S_{i,t}^{(k)} = S_{i,t}^{(k-1)} - \alpha h^{(k)}_{i,t}
\end{equation}
where the learning rate $\alpha$ is fixed to 0.01, which is selected based on our empirical studies.
The initial values of $S$ can be sinusoidal functions if no existing surrogate signals are available, in which case our method is called surrogate-free motion model. When respiration signals are provided, they can be used as the initial surrogates and then optimized by our method, which is denoted as surrogate-optimized motion model.

\subsubsection{Motion Compensated Iterative Reconstruction}
In all the methods above, the reference volume can be initialized with the average of all phase images in 4DCT, which will exhibit quite large artefacts. Once the motion model and surrogate signals are estimated with the method mentioned above, they can be used to update the reference volume by motion compensated iterative reconstruction (MCIR). Briefly, the transformation at each timepoint can be estimated by Eqn.~\ref{eq:motionmodel} and used to compensate for the motion during an iterative reconstruction (similar to the motion compensated super resolution reconstruction performed in \cite{mcclelland2017generalized}, but with reconstructed image having the same resolution as the CT segments), resulting in a motion-free volume if the transformations are accurate.
Multi-resolution approach was used to update the motion model and reference volume alternatively at three levels. At each resolution level, our method would switch between motion model fitting and MCIR for up to six times or until the objective function did not improve. Each run of motion model fitting or MCIR would be terminated when the objective function stopped decreasing or up to 5 iterations of gradient descent had been used.

\section{Experiments}
\subsubsection{Digital Phantom Simulation} \label{sec:eval}
A digital phantom with motion was generated by the 4DXCAT software~\cite{segars20104d} and post-processed by the cid-X software~\cite{eiben2020consistent}. The simulation was controlled by two respiration traces, i.e. the motion of the chest and the diaphragm, which were measured from 2D Cine MRI from a real patient. The diaphragm trace was deliberately delayed by 1 second to add hysteresis, i.e. the breathing would follow different paths during inhalation and exhalation. The chest trace could simulate the skin marker or belt signals normally used to sort 4DCT data in clinical practice. The phantom dataset consisted of images at 182 timepoints, each with size 355$\times$280$\times$115 voxels and resolution of 1$\times$1$\times$3 mm. A movie of the simulated 4DCT can be found in the supplementary material.

Four motion modelling methods were compared, including the groupwise registration method \cite{zhang2013modeling}, surrogate-driven motion model \cite{mcclelland2017generalized}, and our proposed surrogate-free/optimized motion models. The original 4DCT was also included as a baseline. The groupwise registration method alternatively estimated the deformation field sequence and the mid-position image through 5 iterations, using stationary velocity based diffeomorphism \cite{modat2012parametric}.  For 4DCT and groupwise registration, the real-time volumes were obtained by repeating the ten phase images over time. For the surrogate-driven and surrogate-optimized methods, the chest signal and its temporal derivative \cite{mcclelland2017generalized} were used as the fixed or initial surrogate signals, respectively. For the surrogate-free method, the initial surrogate signals were set to the cos and sin functions derived from the respiration phases. Evaluation metrics include average DSC (dice similarity coefficient) between the estimated and the true tumor masks, TRE (target registration error) of the tumor centroid, and the average RMSE (root-mean-square-error) between the estimated and the true image intensities. 

\subsubsection{Real Patient Data}
Our method was also demonstrated on four clinical 4DCT data, two of which (Pat1/Pat2) were from the ROSS-LC clinical trial \cite{price2018results} and the other two (Pat3/Pat4) were from the public TCIA 4D-Lung dataset \cite{balik2013evaluation}. Due to the absence of respiration signals used for 4DCT sorting, the surrogate-driven/optimized motion models could not be applied to these data retrospectively. Therefore, only the surrogate-free motion model was fit on the real patient data, and each slice was treated as an individual CT segment since no timing information was provided to determine which slices were acquired at the same time. Unlike the digital phantom data, there was no ground-truth for real data, so the end-inhalation images estimated by the surrogate-free motion model were visually compared with the end-inhalation phase of 4DCT. To show that our method could estimate irregular breathing motion, all the end-inhlation timepoints were identified according to the optimized surrogate signals and the two timepoints with the deepest and shallowest inhalation among them were selected for visual comparison. 

\section{Results}

\begin{table}[b]
\centering
\setlength{\tabcolsep}{6pt}
\caption{Evaluation metrics in all scenarios. Unit for TRE and RMSE is mm and HU (Houndsfield unit) respectively.}\label{tab1}
\begin{tabularx}{\textwidth}{cccccc}
\hline
Metrics & 4DCT & groupReg & surrDriven & surrFree & surrOptimized \\
\hline
DSC & $0.37\pm0.27$ & $0.46\pm0.17$ & $0.70\pm0.16$ & $0.76\pm0.11$ & $\boldsymbol{0.79\pm0.10}$ \\
TRE & $3.94\pm2.04$ & $3.04\pm1.27$ & $1.17\pm0.76$ & $1.16\pm0.70$ & $\boldsymbol{0.91\pm0.50}$ \\
RMSE & $108.6\pm28.8$ & $107.1\pm31.1$ & $\boldsymbol{40.6\pm13.4}$ & $57.5\pm14.8$ & $\boldsymbol{40.6\pm13.9}$ \\
\hline
\end{tabularx}
\end{table}

\subsubsection{Phantom Experiment Results}
Table \ref{tab1} displays the evaluation metrics for the 4DCT and the four motion models, as explained in section \ref{sec:eval}. The original 4DCT and groupwise registration method have the worst results, highlighting their inability to model the variable motion. The surrogate-free method is comparable to the surrogate-driven method, although the RMSE metric is slightly worse. By leveraging existing respiration signals, the surrogate-optimized method achieved the best results in all the metrics.

\begin{figure}[t]
\includegraphics[width=\textwidth]{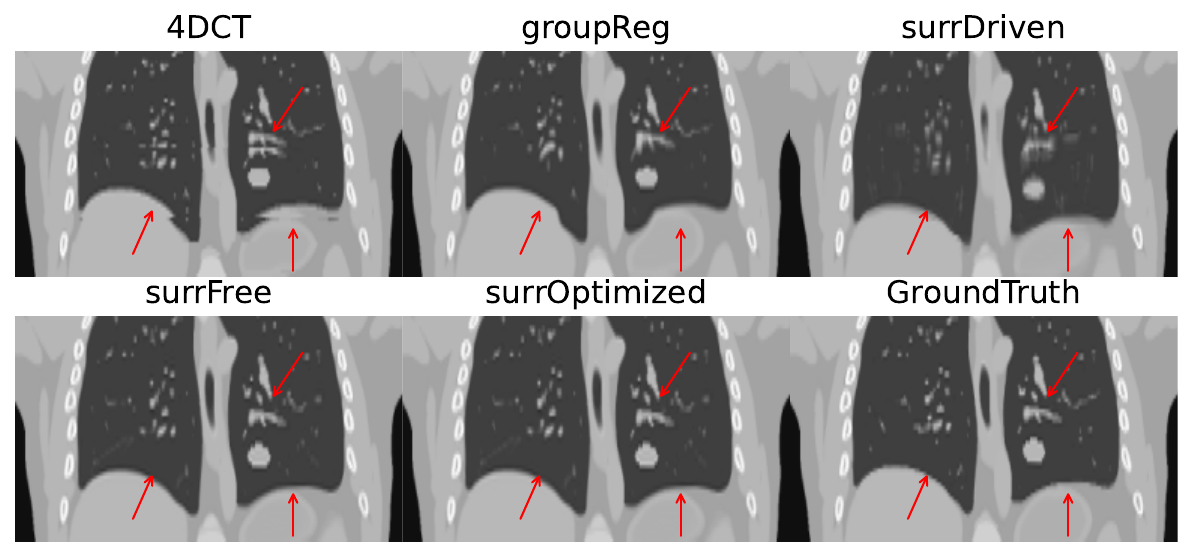}
\caption{ Effects of artefact correction by different methods on digital phantom data. } \label{fig2}
\end{figure}

Fig.~\ref{fig2} shows the end-inhalation phase image of the original 4DCT and the one obtained by groupwise registration, as well as the images obtained by the surrogate-driven/surrogate-free/surrogate-optimized methods, corresponding to the timepoint when the inhalation is deepest. Ground-truth image at the same timepoint as shown by the model is also plotted. While the four motion correction methods all reduce the artefacts in the 4DCT, the proposed surrogate-free/optimized methods obtain the sharpest images without the apparent distortion of the diaphragm and lung tissues, as illustrated by the red arrows. 

\begin{figure}[t]
\includegraphics[width=\textwidth]{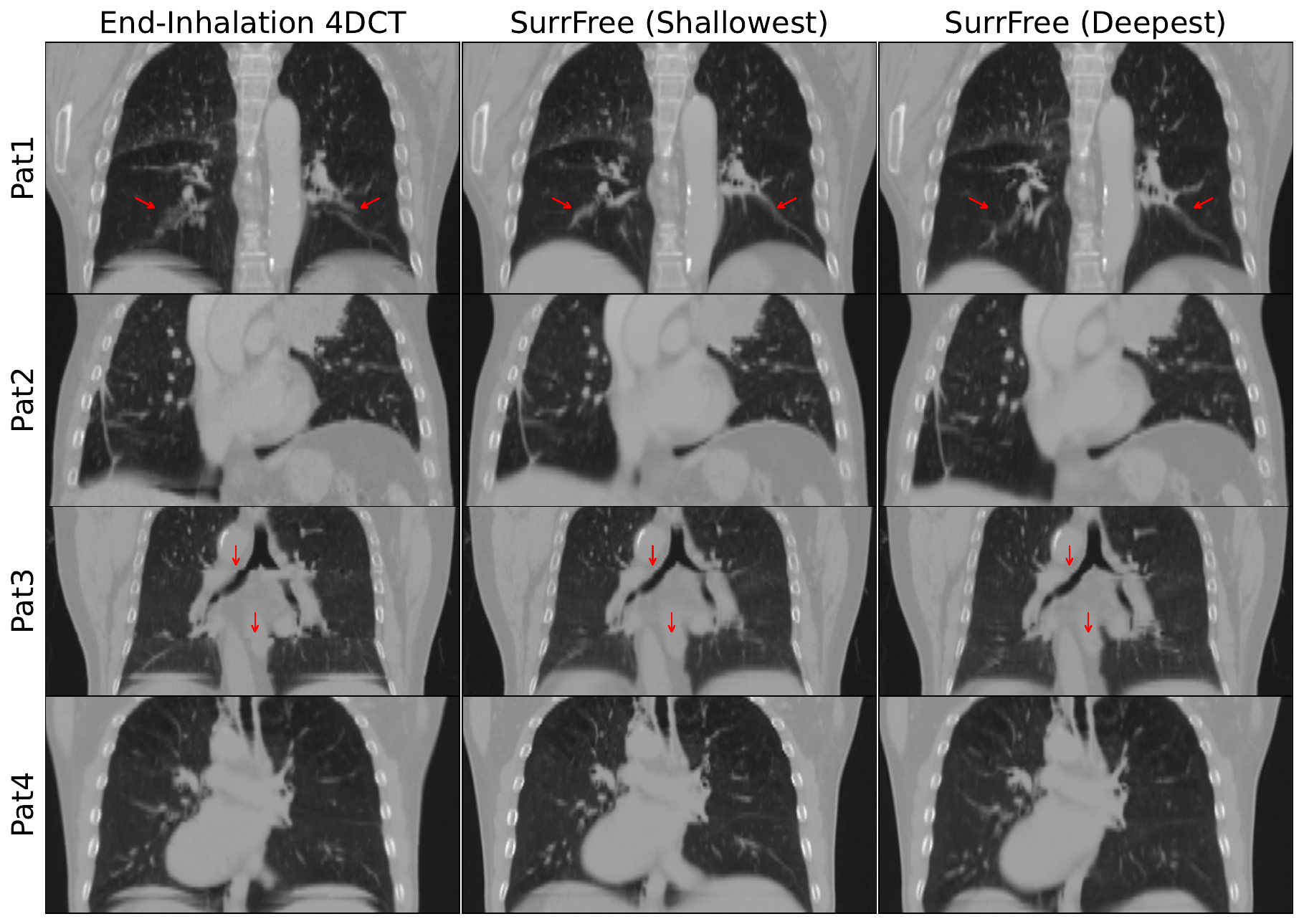}
\caption{ Visual comparison between end-inhalation 4DCT and the deepest/shallowest end-inhalation images estimated by the surrogate-free motion model. } \label{fig3}
\end{figure}

\subsubsection{Real Patient Results}
Fig.~\ref{fig3} shows the end-inhalation 4DCTs (left column), and the shallowest (middle) and the deepest (right) end-inhalation images estimated by the surrogate-free motion model for the four real patients. The sorting artefacts around the diaphragm were reduced in all the patients. For Pat1/3, the duplicate and discontinuity of certain structures were repaired by our method, as indicated by the red arrow. Fig.~\ref{fig3} also showed the variability of the end-inhalation phases due to the irregular breathing, i.e. the depth of end-inhalation could vary across breath cycles as indicated by the difference between the middle and the right column in Fig.~\ref{fig3}. Our method could estimate the variation of breath which 4DCT could not estimate. Movies showing the estimated variable motion for all the patients and the phantom can be found in the supplementary.

\section{Discussion and Conclusion}
Irregular breathing has been a long-standing problem for 4DCT, and the associated artefacts will impact many down-stream applications such as radiotherapy treatment planning, PET imaging and ventilation map derivation. Resolving the irregular breathing motion is challenging due to the limited data available at each slice location, which is only imaged for a single, arbitrary, breath cycle. By applying the surrogate-driven motion model, all the CT segments acquired at different timepoints can contribute to fitting the same set of motion model parameters and forming the motion-free reference image, which can solve the irregular breathing problem if synchronized surrogate signals exist and have strong correlation with internal motion. However, surrogate signals are not always stored after the 4DCT data have been sorted, which hinders retrospective application of surrogate-driven motion model on 4DCT data for clinical verification or research purpose. In addition, the existing surrogate signals often do not have strong correlation with the internal motion.

Our proposed method does not require existing surrogate signals so can be applied to any retrospective 4DCT dataset, including open datasets available online. It encourages revisiting the studies built on top of those dataset to check if the improved 4DCT images can lead to different clinical findings or provide better image-based biomarkers for outcome prediction, which is one of our future research directions. When surrogate signals are available our method can be used to optimise the signals, correcting for weak correlations between the measures signals and the internal motion.

The hypergradient method, although normally used in the meta-learning domain, can be adapted to our own problem and has been found to perform better than alternating the updates to the spatial correspondence models and the surrogate signals, with the other being fixed during the update. With the hypergradient method, most of the calculation of the gradient can be reused for both the gradient for the spatial correspondence models and for the surrogate signals, which can reduce the computation time by approximately a factor of 2. Currently, the execution time for each 4DCT is about 15 to 30 minutes on an Intel Core i7-10700K CPU of 3.8GHz and 8 cores. Acceleration by GPU will be investigated in the future. 

Our method is not limited to 4DCT but can also be applied to other 4D imaging data such as interleaved 4DMRI \cite{eiben2024respiratory} and CBCT \cite{huang2024surrogate}. By reducing the artefacts and more importantly estimating the variable breathing motion, our method can potentially benefit all thoracic cancer patients treated with radiotherapy, as well as other clinical applications that rely on 4DCT.

\begin{credits}
\subsubsection{\ackname} This study was funded by Elekta Ltd., Crawley and the EPSRC-funded UCL Center for Doctoral Training in Intelligent, Integrated Imaging in Healthcare (i4health) (EP/S021930/1). We thank Dr. Gareth Price from the Christie NHS, University of Manchester, UK for providing data from ROSS-LC clinical trial.

\subsubsection{\discintname}
The authors have no competing interests to declare that are relevant to the content of this article.
\end{credits}

%
%
%
\bibliographystyle{splncs04}
\bibliography{refs}
\end{document}